\documentclass[letterpaper]{article} 
\usepackage{aaai2026}  
\usepackage{times}  
\usepackage{helvet}  
\usepackage{courier}  
\usepackage[hyphens]{url}  
\usepackage{graphicx} 
\usepackage{natbib}  
\usepackage{caption} 
\usepackage{algorithm}
\usepackage{algorithmic}

 \usepackage{booktabs}
\usepackage{amsmath}     
\usepackage{amssymb}    
\usepackage{amsfonts}   
\usepackage{mathtools}  
\UseRawInputEncoding

\usepackage{newfloat}
\usepackage{listings}
\DeclareCaptionStyle{ruled}{labelfont=normalfont,labelsep=colon,strut=off} 
\floatstyle{ruled}
\newfloat{listing}{tb}{lst}{}
\floatname{listing}{Listing}
\title{Prompt-Free SAM-Based Multi-Task Framework for Breast Ultrasound Lesion Segmentation and Classification.}

\author{
    Samuel E. Johnny\textsuperscript{\rm 1},
    Bernes L. Atabonfack\textsuperscript{\rm 1},
    Israel Alagbe\textsuperscript{\rm 1},
    Assane Gueye\textsuperscript{\rm 1}
}

\affiliations{

    \textsuperscript{\rm 1}Carnegie Mellon University Africa\\
    Kigali, Rwanda\\
    
}

\usepackage{bibentry}

\begin{document}

\maketitle



\author{Bernes Atabonfack\\
Carnegie Mellon University Africa\\
Kigali, Rwanda\\
\and
Samuel E. Johnny \\
Carnegie Mellon University Africa\\
Kigali, Rwanda\\
\and
Israel Alagbe\\
Carnegie Mellon University Africa\\
Kigali, Rwanda\\
\and
Assane Gueye \\
Carnegie Mellon University Africa\\
Kigali, Rwanda\\
}




\maketitle

\begin{abstract}
    Accurate tumor segmentation and classification in breast ultrasound (BUS) imaging remain challenging due to low contrast, speckle noise, and diverse lesion morphology. This study presents a multi-task deep learning framework that jointly performs lesion segmentation and diagnostic classification using embeddings from the Segment Anything Model (SAM) vision encoder. Unlike prompt-based SAM variants, our approach employs a prompt-free, fully supervised adaptation where high-dimensional SAM features are decoded through either a lightweight convolutional head or a U-Net–inspired decoder for pixel-wise segmentation. The classification branch is enhanced via mask-guided attention, allowing the model to focus on lesion-relevant features while suppressing background artifacts. Experiments on the PRECISE 2025 breast ultrasound dataset, split per class into 80\% training and 20\% testing, show that the proposed method achieves a Dice Similarity Coefficient (DSC) of 0.887 and an accuracy of 92.3\%, ranking among the top entries on the PRECISE challenge leaderboard. These results demonstrate that SAM-based representations, when coupled with segmentation-guided learning, significantly improve both lesion delineation and diagnostic prediction in breast ultrasound imaging.
\end{abstract}

\section{Introduction}

Breast cancer remains the most prevalent malignancy and the leading cause of morbidity and cancer-related mortality, especially among women worldwide, representing a critical public health challenge that requires accurate and timely diagnostic interventions. Early detection is important to improve treatment efficiency, reduce mortality rates, and improve patient quality of life , especially in resource-constrained clinical environments. Among the various imaging modalities used for breast cancer screening, ultrasound imaging has emerged as a particularly valuable tool, offering a noninvasive, radiation-free approach that is especially effective for younger women and individuals with dense breast tissue, where it excels at distinguishing between fluid-filled cysts and solid tumors. However, the interpretation of ultrasound images is often challenged by low contrast, speckle noise, and inter-operator variability, which necessitate automated, robust diagnostic tools.

Recent advances in deep learning (DL) have considerably improved the performance of computer-aided diagnostic (CAD) systems for breast cancer. These systems typically perform two main functions: tumor segmentation, which delineates lesion boundaries, and classification, which determines if the pathology is benign, malignant, or normal. While some systems address these tasks separately, joint multi-task learning (MTL) frameworks are emerging as a powerful solution that allows for shared feature learning and reciprocal task enhancement.

In this study, we propose a novel multi-task framework that jointly performs segmentation and classification of breast ultrasound tumors from a single scan.

\section{Related Work}
Early computer-aided diagnosis (CAD) systems for breast cancer relied on independent pipelines for segmentation and classification, which limited model generalization and performance, especially on small datasets. Dajam and Qahmash \cite{dajam2025breast} introduced a dual-stream architecture combining MobileNet and U-Net to separately address classification and segmentation, achieving 98\% accuracy on the BUSI dataset \cite{sabahesaraki_2018_breast}, though without any shared feature interaction between tasks.

The U-Net architecture \cite{ronneberger2015unet} remains foundational in medical image segmentation due to its encoder–decoder symmetry and effectiveness with limited annotations. Numerous extensions have improved its precision and efficiency. For instance, Dar and Ganivada \cite{dar2024efficientunet} proposed EfficientU-Net, which integrates atrous convolutions and EfficientNet blocks to preserve edge details while reducing computational cost.

More recent research has shifted toward multi-task learning, enabling shared representations between segmentation and classification. Luo et al. \cite{luo2022segmentation} introduced a segmentation-to-classification (S2C) strategy where segmentation-derived attention refines diagnostic prediction. He et al. \cite{he2024multitask} extended this concept through ACSNet, incorporating deformable spatial attention and gated feature transfer to reduce task interference and improve information sharing.

Aumente-Maestro et al. \cite{aumente2025multi} developed an end-to-end multi-task CNN for simultaneous tumor segmentation and classification, emphasizing the role of data quality. They curated a refined subset of the BUSI dataset \cite{sabahesaraki_2018_breast} with 450 consistent images, achieving approximately 15\% performance improvement over single-task baselines (Dice score of  0.75 and classification accuracy of about 80\%). This work underscored the importance of high-quality data and task coupling for robust breast ultrasound analysis.

Wei et al. \cite{wei2024novel} proposed a multi-feature fusion multi-task (MFFMT) network integrating attention modules to exploit shared features between segmentation and classification. Their model achieved 95\% accuracy on BUSI and 87\% on the more challenging MIBUS video dataset \cite{lin2022new}, demonstrating strong generalization across modalities. Similarly, Islam et al. \cite{ISLAM2024100555} introduced EDCNN, an ensemble of Xception \cite{chollet2017xception} and MobileNet \cite{howard2017mobilenetsefficientconvolutionalneural} for classification, and U-Net \cite{10.1007/978-3-319-24574-4_28} for segmentation. Although their approach achieved classification accuracies of 87.82\% and 85.69\% on the BUSI \cite{aldhabyani2020dataset} and UDAIT \cite{8003418} datasets, respectively, it maintained distinct networks for each task rather than a unified multi-task design.

Ensemble and multi-task learning methods have shown promise for breast ultrasound analysis. An ensemble approach combining full image, bounding box, and masked image classifiers with multiple segmentation backbones employed hard voting for malignancy classification and soft voting for BI-RADS categorization \cite{BOBOWICZ2024105522}. A multi-task framework simultaneously addressed segmentation and classification through shared feature extraction with prediction refinement on a curated BUSI dataset \cite{AUMENTEMAESTRO2025108540}. Foundation models like SAM have also been explored, with SAMAug-C using SAM-generated augmentations to boost classification performance \cite{gu2024boostingmedicalimageclassification}, though this approach relies on augmentation rather than direct feature integration and still requires external prompts for segmentation.

Recent advances such as the Segment Anything Model (SAM) \cite{kirillov2023segany} have demonstrated exceptional generalization for object segmentation across diverse domains. However, most SAM-based approaches depend on external prompts (e.g., points, boxes, or masks) and are therefore not directly applicable to fully automated medical workflows, where end-to-end lesion localization and classification are required. To address this limitation, we propose a prompt-free, multi-task SAM adaptation tailored for breast ultrasound analysis. Our framework leverages SAM’s powerful vision encoder to extract rich feature embeddings, which are decoded for lesion segmentation and concurrently used for classification. By integrating a mask-guided attention mechanism, the model explicitly links spatial localization with diagnostic prediction, improving both interpretability and diagnostic accuracy. The following sections describe the architecture and design rationale of the proposed system in detail.

\section{Methodology}

\subsubsection{Motivation}
\label{sec:motivation}

Lesion boundary characteristics are established diagnostic markers in breast imaging. Clinical studies across multiple modalities demonstrate that irregular shape, spiculated margins, and boundary diffuseness significantly correlate with malignancy in ultrasound~\cite{Levman2011Margin}, MRI~\cite{Temerik2023Morphological}, and mammography~\cite{Mohapatra2022Mammography} (P $<$ 0.01). 

Our segmentation-guided architecture is motivated by two objectives:

\begin{itemize}
    \item To explicitly encode diagnostically relevant morphological features, and
    \item To enhance model interpretability by providing spatial localization alongside classification predictions.
\end{itemize}

The predicted segmentation masks offer clinicians a visual explanation of which regions the model identifies as lesions, while the mask-guided attention mechanism ensures classification decisions are grounded in these localized regions rather than spurious image features.

\label{method}
\subsection{Dataset}
\label{dataset}

For this study, we utilized a joint combination of openly available datasets. These include BUSI\cite{aldhabyani2020dataset}, BrEAST \cite{pawlowska2024curated}, and BUS-BRA\cite{gomez2024busbra}. Previous approaches, \cite{aumente2025multi, madhu2024ucapsnet, shilaskar2025classification, pawlowska2024curated}, have used only \cite{aldhabyani2020dataset}, which was the first open-source dataset for breast cancer classification. We used these three datasets to ensure their robustness and increase their distribution. For the classification task, the dataset consists of 1553 Benign, 765 Malignant and 182 Normal class distributions. Segmentation task includes concatenating multiple segmented masks for some data, as it signifies multiple instances of cancer in an individual image. To balance the distribution of the skewed data, we used several data augmentation techniques for the Normal category and a less severe augmentation technique for the Malignant class. Details of the data augmentation and other class balancing information are in the \ref{exp_details} subsection.

\begin{table}[htbp]
\centering
\small
\caption{Breast Cancer Dataset Composition}
\label{tab:training_dataset}
\resizebox{\columnwidth}{!}{%
\begin{tabular}{@{}lccccc@{}}
\toprule
\textbf{Category} & \textbf{BUSI~\cite{aldhabyani2020dataset}} & \textbf{BrEaST~\cite{pawlowska2024curated}} & \textbf{BUS-BRA~\cite{gomez2024busbra}} & \textbf{Images} & \textbf{Masks} \\
\midrule
Benign     & 597 & 99  & 857 & 1,553 & 1,561$^*$ \\
Malignant  & 292 & 63  & 410 & 765   & 765       \\
Normal     & 182 & 0   & 0   & 182   & 182       \\
\midrule
\textbf{Total} & \textbf{1,071} & \textbf{162} & \textbf{1,267} & \textbf{2,500} & \textbf{2,508} \\
\bottomrule
\multicolumn{5}{l}{\scriptsize $^*$BrEaST has 13 extra multi-annotated masks (107 masks for 99 images).} \\
\end{tabular}}
\end{table}

\subsection{Experimental Details}
\label{exp_details}

\subsection{Problem Definition}

Let $\mathcal{X} \in \mathbb{R}^{H \times W \times C}$ denote the input space of breast ultrasound images, where $H$ and $W$ represent the height and width of the image, and $C$ denotes the number of channels. For a given ultrasound image $X \in \mathcal{X}$, our objective is to jointly perform two interconnected tasks: tumor segmentation and tumor classification.

\textbf{Segmentation Task:} The segmentation task aims to generate a pixel-wise binary mask $M \in \{0, 1\}^{H \times W}$, where each pixel $M(i,j)$ indicates whether the pixel at location $(i,j)$ belongs to the tumor region (1) or background (0). Formally, we learn a segmentation function:
\begin{equation}
    f_{\text{seg}}: \mathcal{X} \rightarrow \{0, 1\}^{H \times W}
\end{equation}
that maps an input image $X$ to its corresponding segmentation mask $M$.

\textbf{Classification Task:} Given the input image $X$ and its corresponding segmentation mask $M$, the classification task assigns the tumor to one of $K$ predefined categories. Let $\mathcal{Y} = \{y_1, y_2, \ldots, y_K\}$ represent the set of tumor classes, where in our case $K=3$ corresponding to benign, malignant, and normal. The classification function is defined as:
\begin{equation}
    f_{\text{cls}}: \mathcal{X} \times \{0, 1\}^{H \times W} \rightarrow \mathcal{Y}
\end{equation}
which outputs a class label $y \in \mathcal{Y}$ for the given image and its segmentation.

\textbf{Multi-task Learning Formulation:} Rather than learning two independent models, we propose a unified deep learning framework that jointly optimizes both tasks. Let $\theta$ denote the shared parameters of the encoder network, and $\theta_{\text{seg}}$ and $\theta_{\text{cls}}$ denote the task-specific parameters for segmentation and classification heads, respectively. The overall objective is to learn a joint function:
\begin{equation}
    f_{\theta}: \mathcal{X} \rightarrow \{0, 1\}^{H \times W} \times \mathcal{Y}
\end{equation}
parameterized by $\theta = \{\theta_{\text{shared}}, \theta_{\text{seg}}, \theta_{\text{cls}}\}$, which simultaneously predicts the segmentation mask $M$ and the classification label $y$ for a given input image $X$.

The key advantage of this multi-task formulation is that the segmentation mask provides spatial localization cues that enhance classification accuracy, while the classification objective guides the segmentation network to focus on discriminative tumor regions. This mutual reinforcement between tasks is achieved through shared feature representations learned by the encoder network. 

\subsection{Our proposed Architecture}
\begin{figure*}[htbp]
    \centering
    \includegraphics[width=1.0\linewidth]{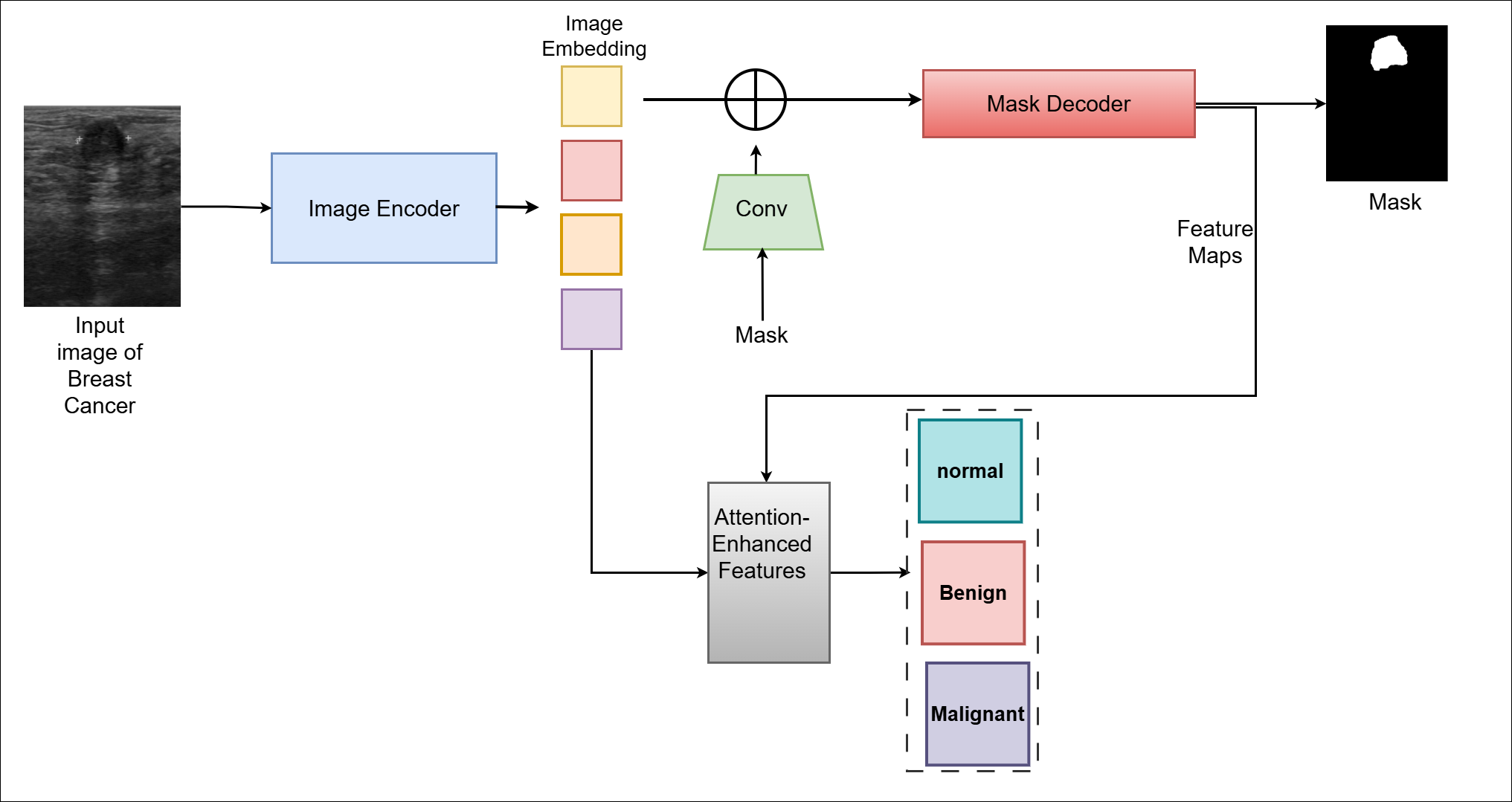}
    \caption{Overview of our proposed multi-task learning framework. The image encoder extracts shared features, which are processed by a mask decoder for segmentation and an attention-enhanced module for classification into benign, malignant, or normal categories.}
    \label{fig:architecture}
\end{figure*}
Figure~\ref{fig:architecture} illustrates the comprehensive architecture of the proposed multi-task learning framework for tumor diagnosis.  The pipeline initiates with a breast ultrasound image, subsequently processed by an image encoder utilizing a Vision Transformer (ViT-B/16) backbone.  The encoder, derived from the Segment Anything Model (SAM)~\cite{kirillov2023segany} and pre-trained on the SA-1B dataset~\cite{kirillov2023segany}, analyzes input images at a resolution of $1024 \times 1024$ pixels to produce multi-scale feature embeddings.  The feature representations function as common inputs for the downstream tasks of segmentation and classification.

 \subsubsection{Segmentation Task}
 
The segmentation branch utilizes high-dimensional feature embeddings extracted from the SAM vision encoder. Unlike the original SAM framework, which uses point or box prompts, our approach learns a direct mapping between breast ultrasound images and corresponding tumor masks.

To reconstruct spatial detail from the encoder output, we experimented with two decoder designs:

A \textit{lightweight convolutional head} composed of stacked convolutional, normalization, and activation layers for efficient upsampling, and

A \textit{U-Net–inspired learned upsampling decoder} that employs transposed convolutions and skip connections to better preserve boundary information.

For cases with multiple annotated masks per image, the model is trained on their combined union, ensuring complete lesion coverage.
Beyond producing pixel-wise delineations, the segmentation branch also provides spatial guidance cues that enhance the classification process by localizing diagnostically relevant regions and suppressing background tissue artifacts common in ultrasound imagery.



\subsubsection{Classification Task}
The classification branch operates on shared encoder features augmented by segmentation-derived attention.
Specifically, the predicted segmentation map is transformed into an attention weight that modulates the encoder features, enabling the classifier to focus on tumor-relevant activations.
A lightweight fully connected head then predicts one of three diagnostic categories — normal, benign, or malignant.
The shared SAM feature embeddings are refined through mask-guided pooling, enabling the network to emphasize tumor-relevant regions before feeding the resulting representation into fully connected layers for class prediction.



\subsubsection{Training Objective}
To enable joint training of both tasks, we employ a weighted multi-task loss function that combines segmentation and classification objectives.

\textbf{Segmentation Loss:} Following the SAM architecture~\cite{kirillov2023segany}, we combine Binary Cross-Entropy (BCE) loss and Dice loss:
\begin{equation}
    \mathcal{L}_{\text{seg}} = \mathcal{L}_{\text{BCE}} + \mathcal{L}_{\text{Dice}}
\end{equation}

The Binary Cross-Entropy loss measures pixel-wise prediction accuracy:
\begin{equation}
    \mathcal{L}_{\text{BCE}} = -\frac{1}{HW}\sum_{i=1}^{H}\sum_{j=1}^{W} \left[y_{ij}\log(\hat{y}_{ij}) + (1-y_{ij})\log(1-\hat{y}_{ij})\right]
\end{equation}
where $y_{ij} \in \{0,1\}$ is the ground truth label (1 for tumor, 0 for background) and $\hat{y}_{ij} \in [0,1]$ is the predicted probability at pixel $(i,j)$ in an $H \times W$ image.

The Dice loss evaluates region overlap between prediction and ground truth:
\begin{equation}
    \mathcal{L}_{\text{Dice}} = 1 - \frac{2\sum_{i,j}y_{ij}\hat{y}_{ij} + \epsilon}{\sum_{i,j}y_{ij} + \sum_{i,j}\hat{y}_{ij} + \epsilon}
\end{equation}
where $\epsilon = 10^{-6}$ the smoothing factor, prevents numerical instability. This loss is particularly effective for handling class imbalance common in medical imaging.

\textbf{Classification Loss:} We use Cross-Entropy loss to measure classification accuracy:
\begin{equation}
    \mathcal{L}_{\text{cls}} = -\sum_{k=1}^{K} y_k \log(\hat{y}_k)
\end{equation}
where $y_k \in \{0,1\}$ is the one-hot encoded ground truth for class $k$, $\hat{y}_k \in [0,1]$ is the predicted probability, and $K=3$ represents our three classes (benign, malignant, normal).

\textbf{Multi-task Loss:} The final objective combines both losses:
\begin{equation}
    \mathcal{L}_{\text{total}} = \lambda \mathcal{L}_{\text{seg}} + (1-\lambda) \mathcal{L}_{\text{cls}}
\end{equation}
where $\lambda = 0.6$ prioritizes segmentation, as accurate tumor localization is essential for reliable classification and clinical interpretation.

\subsection{Experimental Setup}

All models were trained for 60 epochs using the PRECISE 2025 breast ultrasound dataset.
The dataset was split per class into 80\% for training and 20\% for testing, ensuring a balanced distribution across the normal, benign, and malignant categories.
Each model was trained under identical conditions to enable fair comparison.
We evaluated segmentation performance using Dice coefficient and the Binary Cross-Entropy (BCE) loss, while classification performance was assessed through Accuracy and the Cross-Entropy (CE) loss.

All experiments were conducted on Nvidia L40S GPUs using the Adam optimizer with a learning rate of $\alpha = 0.0001$. To facilitate stable multi-task learning, we employed a curriculum training strategy over three phases. In the initial 20 epochs, the mask-guided attention mechanism received ground truth segmentation masks, allowing the classifier to learn from accurate spatial information. During the subsequent 20 epochs, we transitioned to a mixed training regime where both ground truth and predicted masks were randomly sampled with equal probability, enabling gradual adaptation as segmentation quality improved. In the final 20 epochs, only predicted masks were used, ensuring end-to-end dependency on the segmentation branch and testing the model's ability to leverage its own spatial predictions for classification.

\subsection{Comparative Evaluation}

We compare our proposed method against three state-of-the-art approaches: the multi-task baseline \cite{AUMENTEMAESTRO2025108540}, the BI-RADS ensemble method \cite{BOBOWICZ2024105522}, and EDCNN \cite{ISLAM2024100555}. Table~\ref{tab:overall_results} summarizes the performance across classification and segmentation tasks.

\begin{table}[t]
\centering
\caption{Overall performance comparison on breast cancer classification and segmentation tasks.}
\label{tab:overall_results}
\small
\resizebox{\columnwidth}{!}{%
\begin{tabular}{l c c c c}
\toprule
\textbf{Metric} & \textbf{Baseline\cite{AUMENTEMAESTRO2025108540}} & \textbf{BI-RADS\cite{BOBOWICZ2024105522}} & \textbf{EDCNN\cite{ISLAM2024100555}} & \textbf{Ours} \\
\midrule
\multicolumn{5}{l}{\textit{Classification}} \\[2pt]
Accuracy (\%) $\uparrow$ & 71.06 & 89.23 & 80.45 & \textbf{90.7} \\
F1-Score $\uparrow$ & 0.693 & \textbf{0.893} & 0.7646 & 0.887 \\
AUC $\uparrow$ & 0.761 & \textbf{0.964} & 0.9322 & \textbf{0.981} \\
\midrule
\multicolumn{5}{l}{\textit{Segmentation}} \\[2pt]
DSC $\uparrow$ & 0.153 & 0.806 & 0.583 & \textbf{0.887} \\
HD95 (mm) $\downarrow$ & 83.11 & \textbf{30.63} & 45.48 & 60.838 \\
NSD $\uparrow$ & 0.007 & 0.433 & 0.313 & \textbf{0.503} \\
\bottomrule
\end{tabular}%
}
\end{table}

Our method achieves the highest classification accuracy (90.7\%) and AUC (0.981), outperforming the BI-RADS ensemble by 1.47\% and 1.7\%, respectively. For segmentation, our approach attains superior DSC (0.887) and NSD (0.503), indicating accurate lesion localization. While HD95 is higher than BI-RADS, the improved DSC and NSD demonstrate robust boundary delineation critical for clinical assessment. These results validate the effectiveness of our prompt-free multi-task framework in leveraging SAM's feature representations for end-to-end breast ultrasound analysis.


\subsection{Ablation Studies}

To isolate the contribution of each architectural component, we conduct ablation experiments across three model variants that systematically vary the decoder architecture and attention mechanisms.

\textbf{U-Net Head (UNet-H):} Employs a standard U-Net decoder architecture with symmetric skip connections from each encoder layer to its corresponding decoder layer. Features are progressively upsampled using learned transposed convolutions, and skip connections concatenate multi-scale features to preserve spatial details lost during encoding. This design enables precise boundary reconstruction through the integration of both high-level semantic and low-level spatial information.

\textbf{Simpler Decoder Head (SD-H):} Replaces the U-Net decoder with a lightweight alternative consisting of sequential convolutional layers followed by bilinear upsampling. This variant eliminates skip connections entirely, relying solely on the encoder's bottleneck features for mask reconstruction. The architecture reduces computational complexity and parameter count while testing whether skip connections are necessary for adequate segmentation performance in our setting.

\textbf{Mask-Guided Classification with Attention (MGC-A (UNet-H)):} Augments the U-Net decoder with an attention mechanism that explicitly conditions the classification pathway on predicted segmentation masks. Specifically, the attention module weights the global feature representation using spatial information from the segmentation head, allowing the classifier to focus on lesion-relevant regions. This design tests whether explicit spatial guidance improves diagnostic discrimination beyond standard multi-task learning.





\begin{table}[h]
\centering
\caption{Ablation study comparing segmentation and classification performance across model variants.}
\small
\resizebox{\columnwidth}{!}{%
\begin{tabular}{l c c c c c}
\toprule
\textbf{Metric} & \textbf{UNet-H} & \textbf{SD-H} & \textbf{MGC-A (UNet-H)} \\
\midrule
\textbf{Classification} & & & \\
Accuracy $\uparrow$ & 90.7 & 90.7 & \textbf{92.3} \\
CE Loss $\downarrow$ & 0.310 & 0.277 & \textbf{0.227} \\
AUC $\uparrow$ & \textbf{0.981} & \textbf{0.981} & 0.980 \\
\midrule
\textbf{Segmentation} & & & \\
BCE $\downarrow$ & 0.059 & 0.059 & 0.064 \\
DSC $\uparrow$ & 0.882 & \textbf{0.887} & 0.879 \\
HD95 (mm) $\downarrow$ & 64.458 & \textbf{60.838} & 62.990 \\
NSD $\uparrow$ & 0.498 & \textbf{0.503} & 0.488 \\
\bottomrule
\end{tabular}}
\label{tab:ablation_results}
\end{table}


\subsubsection{Discussion}
The ablation results highlight important trade-offs in model design for breast ultrasound segmentation and classification.
Although the U-Net–based decoder was expected to yield the most precise boundary reconstruction, the Simpler Decoder Head (SD-H) achieved slightly higher segmentation metrics (Dice = 0.802, DSC = 0.887, HD95 = 60.8). This suggests that the rich spatial embeddings provided by the SAM vision encoder already capture sufficient contextual detail, allowing a lightweight decoder to focus effectively on mask reconstruction without introducing additional complexity or overfitting.

In contrast, the Mask-Guided Classification (MGC-A) configuration showed the best classification performance (92.3\% accuracy, CE loss = 0.227), confirming that segmentation-derived spatial cues enhance discriminative feature learning. The mask-guided attention mechanism likely encourages the model to focus on lesion-relevant regions, reducing the influence of background artifacts and improving diagnostic prediction.

Overall, the experiments demonstrate that decoder simplicity and attention-guided feature coupling play complementary roles. The SAM encoder generalizes well even with minimal decoder capacity, while attention-based feature modulation provides measurable gains in classification. These results reinforce the viability of prompt-free SAM adaptations for ultrasound imaging and suggest that efficient decoder designs can achieve strong performance without compromising interpretability or computational cost.

\section{Conclusion}
This work presents a prompt-free, multi-task adaptation of the Segment Anything Model (SAM) for breast ultrasound lesion segmentation and classification.
By leveraging SAM’s vision encoder as a powerful feature extractor and integrating both lightweight and U-Net–style decoders, the framework achieves high segmentation precision and robust diagnostic accuracy without relying on external prompts.
Experimental results on the PRECISE 2025 dataset demonstrate that a simpler convolutional decoder can effectively harness SAM’s rich embeddings, outperforming more complex architectures in segmentation metrics, while mask-guided attention substantially enhances classification performance.

These findings highlight that efficient architectural design and task coupling can yield competitive results even under limited data conditions, suggesting practical applicability for real-world ultrasound workflows.
Future work will explore extending this framework to multi-view and temporal ultrasound data, incorporating uncertainty estimation for clinical reliability, and adapting the prompt-free SAM encoder to other imaging modalities such as MRI.


\bibliography{aaai2026}

\end{document}